\documentclass{article}

\usepackage[preprint]{stylefile}

\makeatletter
\renewenvironment{abstract}{%
  \par\addvspace{0.5em}%
  \centerline{\large\bf Abstract}%
  \vspace{0.5ex}%
}{\par\addvspace{1ex}}
\makeatother

\makeatletter
\renewcommand{\@notice}{}
\makeatother

\usepackage[utf8]{inputenc}
\usepackage[T1]{fontenc}
\usepackage{hyperref}
\usepackage{url}
\usepackage{booktabs}
\usepackage{amsfonts}
\usepackage{amsmath}
\usepackage{colortbl}
\usepackage{amssymb}
\usepackage{nicefrac}
\usepackage{microtype}
\usepackage{float}
\usepackage{xcolor}
\usepackage{graphicx}
\usepackage{placeins}
\usepackage{xspace}

\newcommand{\method}{\textbf{GridProbe}\xspace}

\title{GridProbe: Posterior-Probing for Adaptive Test-Time Compute in Long-Video VLMs}

\author{%
\makebox[\textwidth][c]{%
 \begin{tabular}{@{}c c c c c@{}}
  Mohamed Eltahir$^{1}$ &
  Lama Ayash$^{1}$ &
  Ali Habibullah$^{1}$ &
  Tanveer Hussain$^{2}$\footnotemark[1] &
  Naeemullah Khan$^{1}$\footnotemark[3] 
  \end{tabular}%
}%
\\
$^{1}$ King Abdullah University of Science and Technology (KAUST), Thuwal, Saudi Arabia \\
$^{2}$ Department of Computer Science, Edge Hill University, Ormskirk, England \\
\texttt{\{mohamed.hamid, lama.ayash, ali.habibullah\}@kaust.edu.sa} \\
\texttt{hussaint@edgehill.ac.uk, naeemullah.khan@kaust.edu.sa}}

\begin{document}
\addtolength{\topmargin}{-1.2cm}
\addtolength{\textheight}{0cm}
\maketitle
\vspace{-10mm}
\begin{figure}[H]
  \centering
  \includegraphics[width=\linewidth]{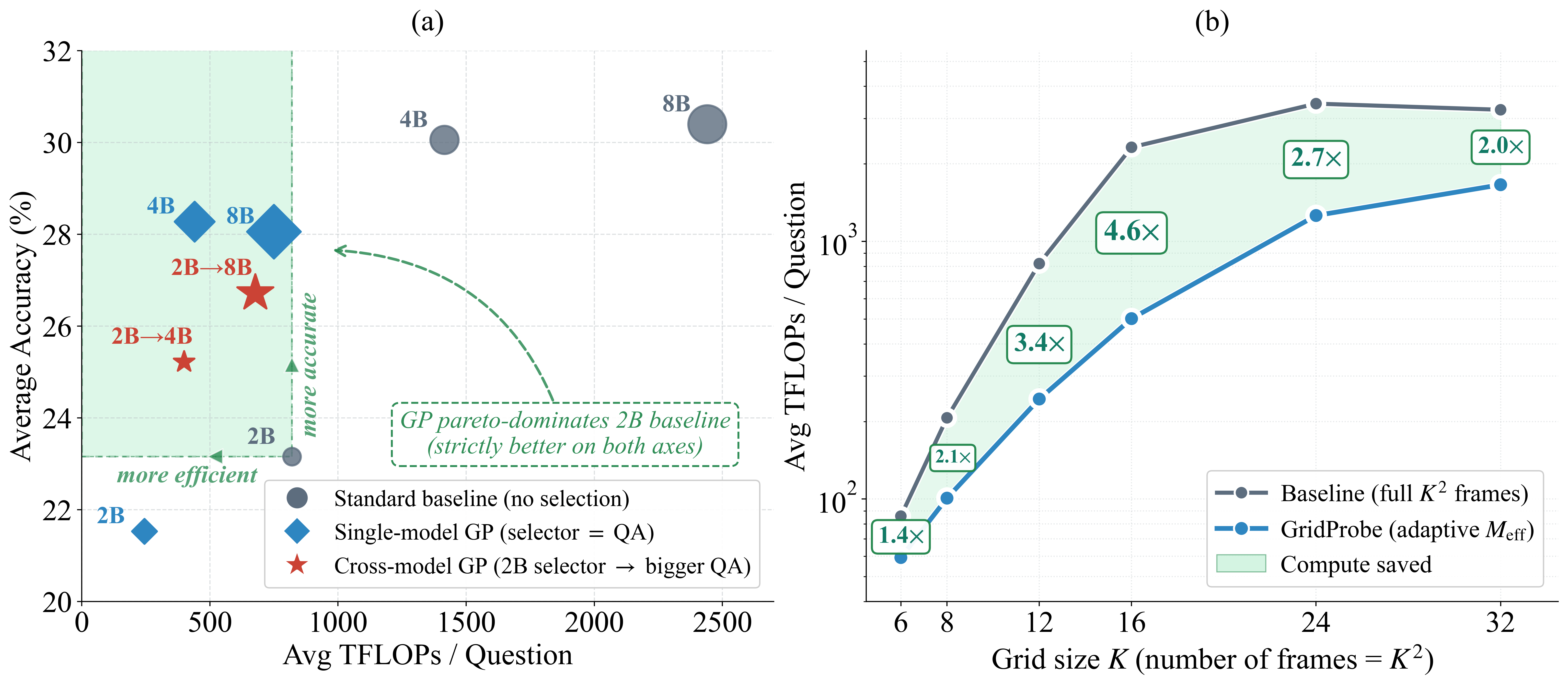}
  \vspace{-6mm}
\caption{\textbf{(a) VMME-V2 Pareto across QA model sizes.} \method{} variants in the green region Pareto-dominate the 2B baseline. \textbf{(b) Compute reduction across $K$ at fixed 2B QA.}}
\label{fig:pareto}
\end{figure}
{
  \renewcommand{\thefootnote}{\fnsymbol{footnote}}
  \footnotetext[1]{Corresponding Author}
  \footnotetext[3]{Principal Investigator (PI)}
  \footnotetext{Code: \url{https://www.github.com/mohammad2012191/GridProbe}}
}

\enlargethispage{2cm}  
\begin{abstract}
Long-video understanding in VLMs is bottlenecked by a single monolithic forward pass over thousands of frames at quadratic attention cost. A common mitigation is to first \emph{select} a small subset of informative frames before the forward pass; common for training-free selectors via auxiliary \emph{encoder-space} similarities. Such signals are capped by contrastive pretraining, which usually fails on reasoning-heavy queries (negation, cross-frame counting, holistic summarization). We propose \emph{\method{}}, an efficient training-free \emph{posterior-probing} inference paradigm that scores evidence in \emph{answer space} using a frozen VLM's own reasoning and then selects question-relevant frames adaptively, resulting in sub-quadratic attention cost with little to no accuracy loss. We arrange frames on a $K{\times}K$ grid and run lightweight row \emph{R} and column \emph{C} probes, where each probe reads its peak posterior as a query-conditioned confidence. The outer product of \emph{R} and \emph{C} yields an interpretable \emph{importance map} whose skewness and kurtosis drive \emph{Shape-Adaptive Selection}, a closed-form rule that reliably replaces the fixed frame budget $M$ with a per-question $M_{\mathrm{eff}}$. We show empirically that $M_{\mathrm{eff}}$, surprisingly, tracks intrinsic question difficulty without ever seeing the answer, a sign of test-time adaptive compute. On Video-MME-v2, \method{} matches the monolithic baseline within $1.6$ pp Avg Acc at $3.36\times$ TFLOPs reduction, while on LongVideoBench it Pareto-dominates the baseline ($+0.9$ pp at $0.35\times$ compute). Because the selector and QA models can be decoupled, pairing a small 2B selector with a stronger 4B or 8B QA is strictly Pareto-dominant over the 2B monolithic baseline (up to $+4.0$ pp at $0.52\times$ compute, on average), with no retraining. Finally, the interpretability of the importance maps opens future avenues for behavioral diagnostics, grounding, and frame-selection
distillation.
\end{abstract}

\section{Introduction}\label{sec:intro}

\vspace{-2mm}
Modern video VLMs process long videos by compressing many frames into one forward pass. Qwen3-VL-2B ~\cite{bai2025qwen3}, for example, uses an adaptive per-frame resolution that crushes individual frames to $\approx 240$ visual tokens when 2048 frames are passed, an order of magnitude below the $\approx 2960$ tokens per frame at the 64-frame setting. This trade-off exchanges per-frame fidelity for temporal coverage and reflects a structural limit: per-token cost is dominated by the (linear-in-tokens) FFN at current scales while attention adds asymptotically quadratic-in-sequence-length growth on top, so reducing the number of input tokens delivers the strongest compute savings, and even models trained with 256K-token contexts cannot afford dense sampling \emph{and} dense attention at scale.

An orthogonal response is \emph{frame selection}: pick the $M \ll N$ most informative frames and run the VLM on only those. Recent training-free selectors (MDP3~\cite{sun2025mdp3}, CLIP-matching, SigLIP-based scoring) and learned variants (Frame-Voyager~\cite{yu2024frame}, Focus~\cite{zhu2025focus}, HFS~\cite{yang2025hfs}) share a common structure: frames and the query are embedded by separate vision and text encoders, and a similarity function in that shared space scores each frame. We call this paradigm \textbf{encoder-space selection}. Its weakness is documented: MDP3's own qualitative analysis shows SigLIP-matching failing on negation, cross-frame counting, and summarization queries, because these queries typically require reasoning outside the encoder's representational capacity.

We argue for a stronger move than swapping in a better selector. The VLM already knows which frames matter, it just needs to be asked. If we feed the VLM a subset of frames with the query, its posterior over the answer space encodes how confidently it can answer \emph{given that subset} (Figure~\ref{fig:answervsencoder}). High confidence on a small subset implies those frames carry the answer. This observation motivates a different inference paradigm rather than a different selector.

\begin{figure}[t]
  \centering
  \includegraphics[width=\linewidth]{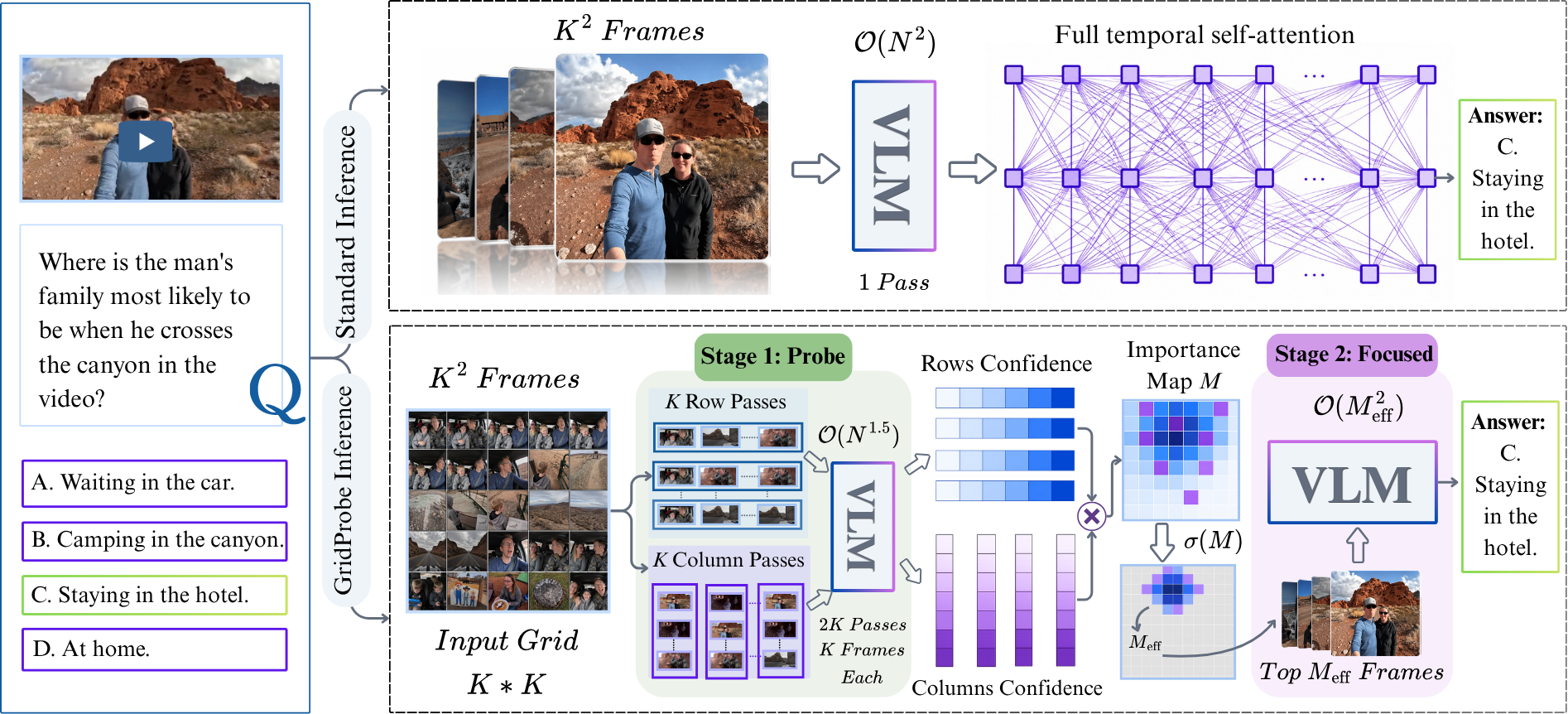}
  \caption{\method{} pipeline. Stage~1: $2K$ row/column probes on $K^2$ candidate frames yield an importance map. Stage~2: one focused pass on the top-$M_{\mathrm{eff}}$ cells, sized adaptively from the map's distribution shape.}
  
  \label{fig:pipeline}
\end{figure}

To this end, we introduce \textbf{\method{}} (Figure~\ref{fig:pipeline}), a training-free \emph{posterior-probing} inference paradigm that replaces the standard one-shot forward pass with a self-probing recipe. We factorize the candidate frame pool into a $K \times K$ grid and run lightweight, axis-aligned probe passes over the rows and columns through a frozen VLM. The outer product of the row and column peak-posterior confidences yields a question-conditioned importance map. By default, the same frozen VLM serves as both the selector and the answerer. We further show that the two roles can be decoupled for a strict Pareto improvement.

This single design shift has three structural consequences. First, the selection signal is \emph{reasoning-grounded}, it inherits the VLM's full reasoning capacity, so negation, cross-frame counting, and compositional queries are handled natively rather than being lost in contrastive embedding. Second, the signal scales with backbone capability without retraining, a stronger VLM automatically yields a sharper importance map. Third, the maps are mechanically interpretable, rendering the model's evidence-gathering legible at the frame level. Notably, the current formulation reads a peak posterior over a finite answer space.

Once frames are scored, selectors must determine how many to pass to the final model. Existing methods enforce a static budget $M$, creating an unavoidable trade-off: they waste compute on highly localized questions and bottleneck accuracy on holistic ones by discarding necessary context. Crucially, the \method{} importance map resolves this natively. We demonstrate empirically that the shape of this importance distribution strongly correlates with question difficulty (Figure~\ref{fig:qmaps}, right). Rather than using a static frame budget, we utilize this insight to introduce \textbf{shape-driven adaptive test-time compute}, which sets the per-question size $M_{\mathrm{eff}}$ via a closed-form rule on the map's skewness and kurtosis.

Coupling answer-space probing with shape-driven adaptive selection yields \method{}, a training-free posterior-probing inference paradigm for long-video VLMs. Three findings anchor our empirical claims: (a) Pareto-dominant cross-model composition without retraining, (b) Pareto-efficient single-model operation, and (c) Adaptive test-time compute mirrors intrinsic difficulty.

\textbf{Contributions:}

\textbf{Posterior-probing inference paradigm.} We formalize \method{}, a sub-quadratic training-free inference method for long-video VLMs that operates in answer space rather than encoder space, replacing the standard one-shot forward pass.

\textbf{Question-conditioned importance map.} A per-question, frame-level importance map exposes the VLM's evidence-gathering for each query, making long-video understanding interpretable.

\textbf{Shape-driven adaptive test-time compute.} A closed-form statistic on the importance map distribution replaces the fixed frame budget $M$ with a per-question $M_{\mathrm{eff}}$ that adapts to the question difficulty.

\textbf{The Redundancy Principle.} Positive-skew (sparse peaks) and negative-skew (redundant high-importance) maps are different distribution shapes that share the same selection answer.

\section{Related Work}\label{sec:related}
\textbf{Long-video VLMs and the cost of monolithic inference.} 
Recent video VLMs such as Qwen3-VL~\cite{bai2025qwen3}, InternVL3.5~\cite{wang2025internvl3}, and LLaVA-Video~\cite{li2025llava} scale to thousands of frames via extended context windows combined with adaptive per-frame visual-token budgets. Despite design differences, all share a structural commitment to a single monolithic forward pass with quadratic attention in input length~$O(N^2)$. Even 256K-token contexts cannot afford dense attention over dense sampling, so reducing the cost of this single forward at inference time, without retraining the backbone or compromising visual fidelity, has become a practical priority. 

\textbf{Encoder-space frame selection.}
A dominant mitigation is to score and select a subset of informative frames before the forward pass. Training-free methods rely on similarities in vision-language encoder space (CLIP~\cite{radford2021learning}, SigLIP~\cite{zhai2023sigmoid}). FOCUS~\cite{zhu2025focus} adds adaptive exploration over this signal, while MDP3~\cite{sun2025mdp3} generalizes ranking into a list-wise subset optimization that captures query relevance, diversity, and sequential structure. Learned variants (Frame-Voyager~\cite{yu2024frame}, HFS~\cite{yang2025hfs}) train auxiliary scoring heads or fine-tune the backbone to emit selection signals, trading training complexity for accuracy. We collectively call this family \emph{encoder-space selection}: the selection signal is computed in a representation space structurally separate from the QA model's reasoning, and its quality is therefore bounded by what that space was trained to encode. Reasoning-heavy queries (negation, cross-frame counting, holistic summarization) routinely defeat encoder-space signals that the QA model itself could resolve natively.

\textbf{Multimodal frame scoring and the static-budget assumption.}
Recent work pushes scoring closer to the QA model. FRAG~\cite{huang2025frag} evaluates each frame with a multimodal model and selects the top-$M$, which moves the signal from encoder space to model space but remains frame-wise (no temporal context, no reasoning about evidence sufficiency). Independently of the scoring axis, prior frame-selection methods share a second assumption: the selection size $M$ is fixed a priori, wasting compute on localized queries (where $M{\ll}K^2$ suffices) and starving holistic queries (where the answer is genuinely dispersed). A scoring signal that captures sub-frame reasoning and a per-question budget that adapts to the shape of the evidence both remain open.

\textbf{Test-time compute and agentic video inference.}
A growing body of work allocates \emph{test-time compute} adaptively to improve answer quality. Text-domain efforts include longer chain-of-thought, self-consistency, and search-based decoding~\cite{guo2025deepseek}. In the video domain, the closest prior work uses LLM-based agents to route compute per question. VideoAtlas~\cite{eltahir2026videoatlas} represents a video as a hierarchical grid explored by a Master-Worker agent loop, achieving logarithmic compute growth with video duration. VideoAgent~\cite{wang2024videoagent} and AVUA~\cite{jeoung2024adaptive} similarly use LLM agents that recursively re-sample frames based on their own intermediate reasoning. These systems achieve adaptive per-question compute by orchestrating multi-step agent loops. They inherit the orchestration overhead, control-flow complexity, and per-question planning costs of multi-step inference. A non-iterative, fixed-schedule mechanism that delivers comparable adaptive-compute behavior without agent orchestration is absent from this line of work.

Three threads converge on the same problem from different angles, each leaving a complementary gap. Encoder-space frame selection decreases input volume but operates in a representation space disconnected from the QA model's reasoning. Multimodal frame scoring bridges to model space but stays frame-wise and locks $M$ a priori. Agentic adaptive inference routes per-question compute through multi-step agent orchestration. What is missing across all three threads is a \emph{fixed-schedule training-free} mechanism that scores in the QA model's own answer space, captures cross-frame reasoning rather than per-frame similarity, and sizes the per-question budget in closed form. We describe how \method{} fills all these gaps in the next section.

\section{Methodology: \method{}}\label{sec:method}

\begin{figure}[t]
  \centering
  \includegraphics[width=0.92\linewidth]{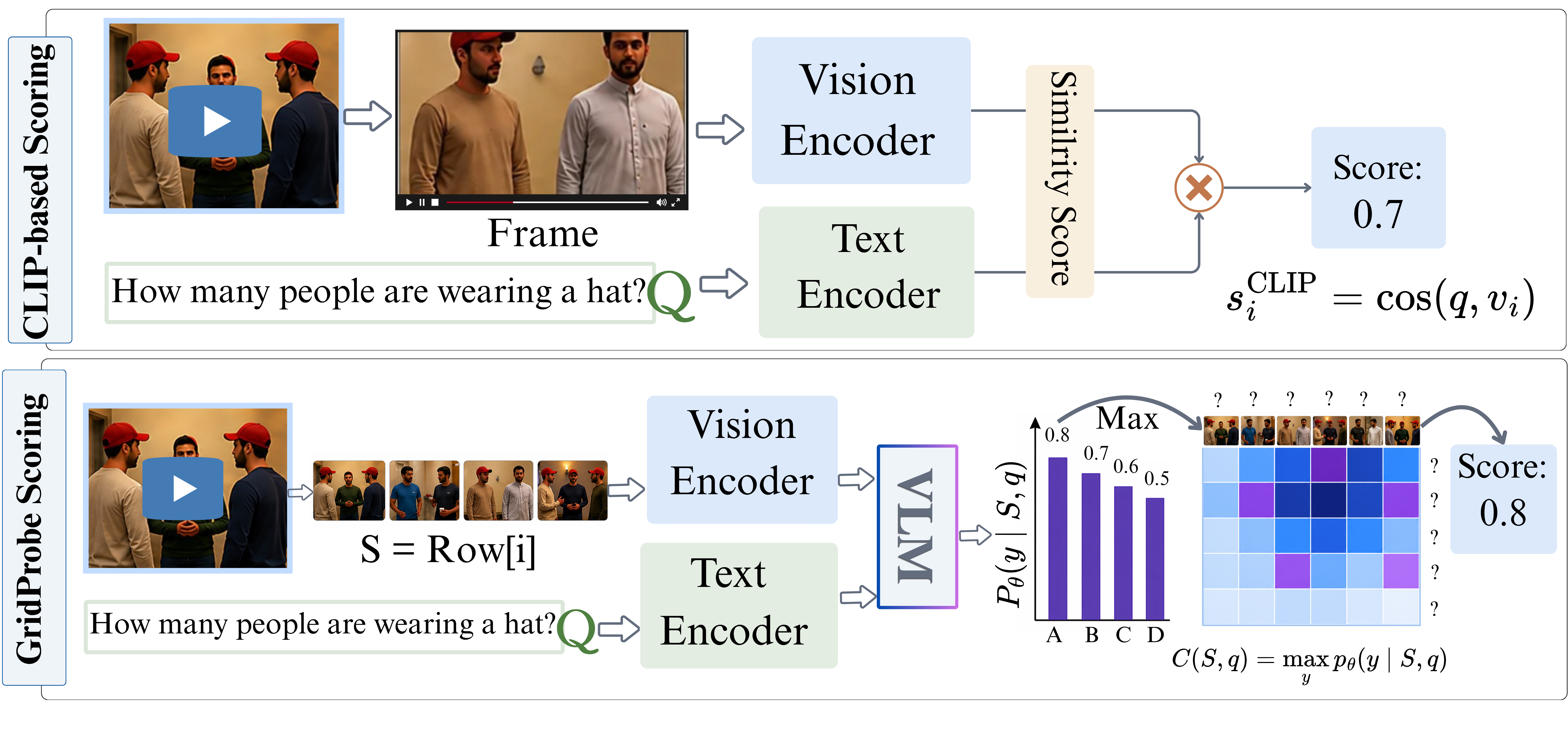}
  \caption{Encoder-space (top) vs answer-space (bottom) selection signals. Encoder-space scoring computes scalar similarity from independent vision and text encoders, while answer-space scoring reads the probe confidence directly from the QA VLM's posterior over answer candidates.}
  \label{fig:answervsencoder}
\end{figure}

\subsection{Setup and Notation}

Let $V = \{f_0, \ldots, f_{n-1}\}$ be an ordered sequence of video frames and $q$ a natural-language query. The answer space $\mathcal{Y}$ depends on the task (for multiple-choice, $|\mathcal{Y}| = 4$). A frozen VLM $\theta$ defines a conditional probability distribution $p_\theta(\cdot \mid S, q)$ over $\mathcal{Y}$ for any frame subset $S \subseteq V$ paired with $q$. We define the \emph{probe confidence} ($c$) as the peak of this posterior: 
\begin{equation}
c(S, q) \;=\; \max_{y \in \mathcal{Y}}\; p_\theta(y \mid S, q).
\label{eq:conf}
\end{equation}
Intuitively, $c(S, q)$ measures how confidently the model can commit to a single answer given $S$. We use this as a proxy for relevance: high confidence implies $S$ contains frames needed to answer $q$, while a flat posterior signals that the subset lacks the evidence to discriminate among the candidates. Figure~\ref{fig:answervsencoder} contrasts this answer-space signal with encoder-space selection, where the score is a similarity computed by independent vision and text encoders.

\subsection{Grid Formulation and Importance Map}

We sample $K^2$ frames uniformly from $V$ and index them as a conceptual $K\times K$ grid. For each row $r \in \{0,\ldots,K-1\}$ and column $c \in \{0,\ldots,K-1\}$, we define
\begin{equation}
S^{\mathrm{row}}_r \;=\; \{\,f_{rK + j}\,\}_{j=0}^{K-1},
\qquad
S^{\mathrm{col}}_c \;=\; \{\,f_{c + jK}\,\}_{j=0}^{K-1}.
\end{equation}
giving $K$ row subsets (local temporal coverage) and $K$ column subsets (strided, periodic coverage). In total, $2K$ probe passes are required, each seeing only $K$ frames.

The $K$ row subsets provide \textbf{local temporal coverage}: each row groups $K$ contiguous frames from a localized segment of the video timeline, exposing fine-grained event-local evidence. The $K$ column subsets provide \textbf{strided periodic coverage}: each column groups $K$ frames at stride $K$, sampling the full timeline at uniform intervals and exposing distributed or recurring evidence. The two axes are complementary: any grid cell $(r, c)$ is uniquely indexed by the intersection of one local row and one global column, so the same frame is scored once from a local-context view and once from a global-context view. Prior multimodal frame scoring~\cite{huang2025frag} computes per-frame evidence one frame at a time, requiring $K^2$ forward passes to score all $K^2$ candidates. Our row$+$column factorization recovers a cell-level importance map at only $2K$ axis-level forward passes, each seeing only $K$ frames.


For each axis subset we compute the probe confidence via Eq.~\ref{eq:conf}: $c^{\mathrm{row}}_r = c(S^{\mathrm{row}}_r, q)$ and $c^{\mathrm{col}}_c = c(S^{\mathrm{col}}_c, q)$.
The joint importance ($M$) of the grid cell $(r, c)$, corresponding to frame $f_{rK+c}$, is the product
\begin{equation}
M[r, c] \;=\; c^{\mathrm{row}}_r \cdot c^{\mathrm{col}}_c.
\label{eq:map}
\end{equation}
Intuitively, a cell is important only if \emph{both} the row and the column containing it produce confident answers (regardless of whether they are correct or not, as high confidence indicates relevance, not correctness). If only one marginal is confident, the cell is assigned moderate weight (partial evidence) and downweighted if neither.

In summary, the grid factorization combines local and strided periodic coverage in a single $O(K)$-pass scoring stage and produces a cell-level question-conditioned importance map without the per-frame scoring overhead of prior multimodal-scoring approaches.

\subsection{Adaptive Selection Size via Distribution-Shape Statistics}\label{sec:selector}

Given the $K\times K$ importance map $M$, we need to pick how many cells to keep. A static $M_{\mathrm{eff}}$ is suboptimal: holistic questions benefit from many frames while localization queries need only a few. Our central observation is that these question types leave distinct fingerprints on $M$ itself. A localization query concentrates evidence in a few cells, producing a sharply peaked, right-skewed map. A redundancy-heavy query spreads high importance across many overlapping cells, producing a left-skewed map. A holistic query distributes evidence broadly but sparsely, producing a near-uniform map. Across question types the shape of $M$ co-varies with how hard the question is to answer from few frames, so we hypothesize that \emph{distribution shape itself is an indirect signal for the optimal selection size $M_{\mathrm{eff}}$}.

To act on this hypothesis we capture shape with two complementary moments combined into a single statistic $\sigma(M)$ that drives the adaptive size:
\begin{equation}
\sigma(M) \;=\; \big|\, \mathrm{skew}(M) \,\big| \;+\; 0.5 \cdot \max\!\big(0,\, \mathrm{kurt}_{\text{ex}}(M)\big),
\qquad
M_{\mathrm{eff}} \;=\; \left\lceil \frac{K^{2}}{1 + \gamma_0 \cdot K \cdot \sigma(M)} \right\rceil.
\label{eq:selector}
\end{equation}
Here $\mathrm{skew}(\cdot)$ is the third standardized moment (asymmetric concentration of evidence) and $\mathrm{kurt}_{\text{ex}}(\cdot)$ is the excess fourth standardized moment (peakedness). Each captures a complementary departure from uniformity. Skewness detects evidence biased toward a small subset of cells. Excess kurtosis detects sharp peaks even in symmetric distributions. On a perfectly uniform map the sample variance is zero and the standardized moments are formally undefined; we set $\sigma{=}0$ in this degenerate case (implemented numerically via a variance threshold), so $M_{\mathrm{eff}} = K^2$ and the method falls back to the full pool, equivalent to the monolithic baseline. On a one-hot map ($\sigma \to \infty$ formally), $M_{\mathrm{eff}} \to 1$. In practice $M_{\mathrm{eff}}$ varies smoothly between these extremes per question. The half-weight on kurtosis downweights its larger absolute scale relative to skewness. The factor of $K$ in the denominator (rather than just $\gamma_0 \sigma$) keeps $M_{\mathrm{eff}}$ growing linearly with $K$ on peaked maps instead of quadratically. Without it, doubling $K$ to gain finer probe resolution would also quadruple $M_{\mathrm{eff}}$ on the same map, undoing the focused-pass savings. 

\S\ref{sec:adaptive_compute} validates this distribution shape hypothesis empirically and shows how $\sigma(M)$ helps to allocate more compute exactly to the questions the QA model finds intrinsically hardest.

\begin{figure}[t]
  \centering
  \includegraphics[width=0.75\linewidth]{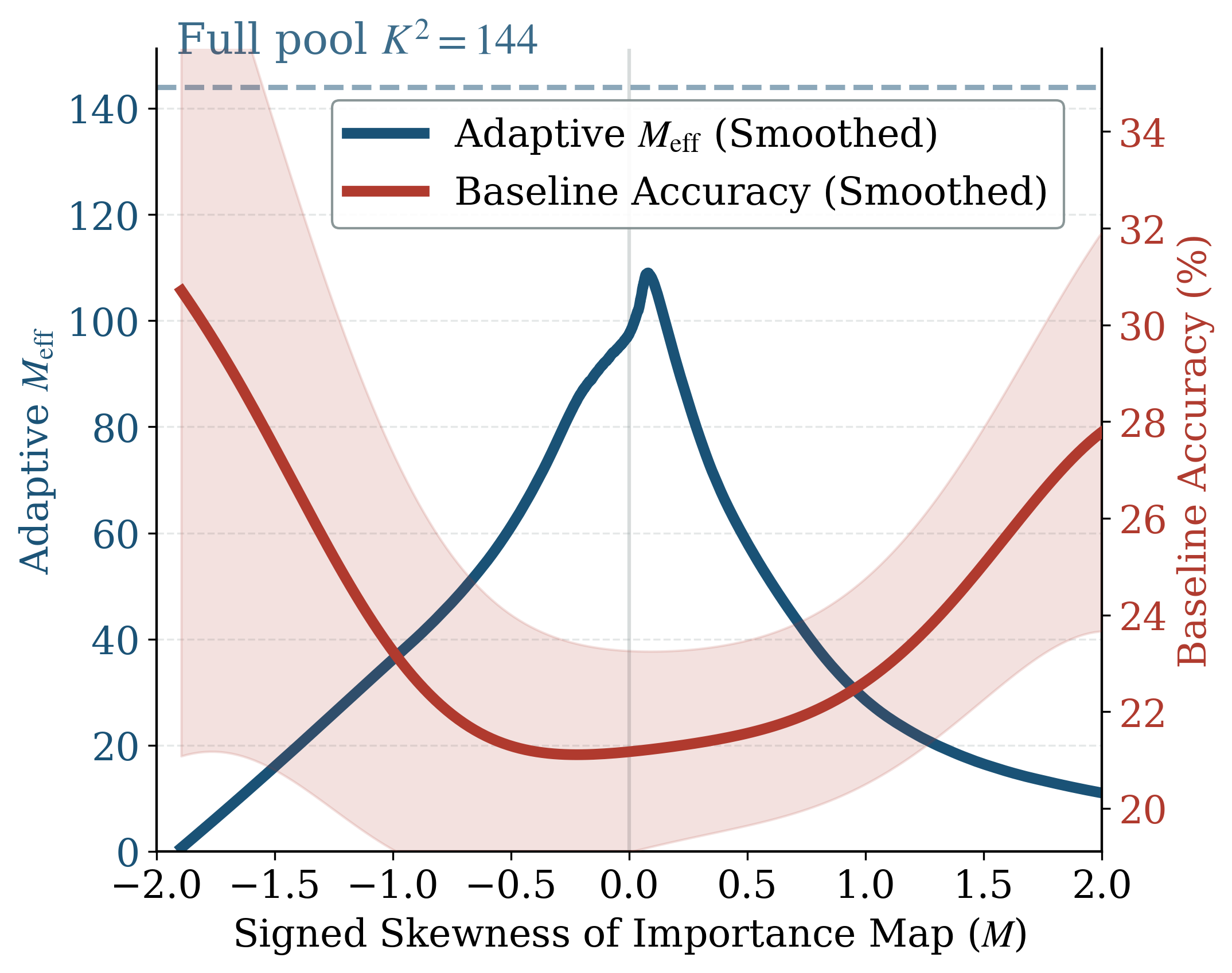}
  \caption{\method{}'s adaptive $M_{\mathrm{eff}}$ (blue) and the 2B baseline accuracy (red), smoothed across signed $\mathrm{skew}(M)$ on V2 ($K{=}12$, $n{=}3{,}200$). The two curves mirror each other: both signed extremes route to small $M_{\mathrm{eff}}$ on intrinsically easier questions, while the near-uniform middle gets near-$K^2$ coverage on intrinsically harder ones, an empirical realization of the redundancy principle (\S\ref{sec:selector}).}
  \label{fig:skew}
\end{figure}

\paragraph{Why $|\mathrm{skew}|$? The redundancy principle.} The absolute value collapses two regimes that have opposite distribution geometries but \emph{identical} selection requirements. A right-skewed map (positive skew, mass at low importance) is the sparse-peak regime, where a few decisive frames carry the answer and the rest can be discarded. A left-skewed map (negative skew, mass at high importance) is the redundancy regime, where most frames are individually informative for the query but show overlapping content, so a small representative subset suffices. The truly compute-hungry case is the low $|\mathrm{skew}|$ near-uniform map, where evidence is sparse-and-dispersed across the timeline and full coverage is warranted. 

Figure~\ref{fig:skew} makes the inverted-U pattern in $M_{\mathrm{eff}}$ explicit: both signed extremes of $\mathrm{skew}(M)$ route to small $M_{\mathrm{eff}}$ while only the near-uniform middle draws near $K^2$ coverage, confirming that $|\mathrm{skew}|$ correctly groups the two ``few-needed'' regimes together. Figure~\ref{fig:qmaps} realizes the three regimes qualitatively on Video-MME-v2 clips, where questions produce $M_{\mathrm{eff}}$ from $140$ (holistic) to $5$ (specific).

\subsection{Two-Stage Inference Pipeline}

\method{} (Figure~\ref{fig:pipeline}) combines the probe and a focused pass:
\begin{enumerate}
\item \textbf{Stage 1 (probe):} run $K$ row-passes and $K$ column-passes on $K$-frame subsets through the frozen VLM. Record $2K$ probe confidences and build $M$ via Eq.~\ref{eq:map}.
\item \textbf{Compute} $M_{\mathrm{eff}}$ from the shape statistic in Eq.~\ref{eq:selector}.
\item \textbf{Stage 2 (focused pass):} select the $M_{\mathrm{eff}}$ frames corresponding to the top entries of $M$, denoted $S^\star$. Run the VLM once on $S^\star$ at full resolution and read off the final answer as $\arg\max_y p_\theta(y \mid S^\star, q)$. %
\end{enumerate}

\subsection{Complexity}
A monolithic pass on $N{=}K^2$ frames has attention cost $O(N^2)$ in the attention-dominated regime. \method{} runs $2K$ probe passes of $K$ frames each (cost $O(K \cdot K^2) = O(N^{1.5})$) plus one focused pass of $M_{\mathrm{eff}}$ frames. For non-uniform importance maps $M_{\mathrm{eff}} \ll N$ and the total attention cost is $O(N^{1.5} + M_{\mathrm{eff}}^2)$, sub-quadratic. In the worst case (perfectly uniform maps) $M_{\mathrm{eff}}{\to}N$ and the focused pass falls back to the monolithic baseline. Empirically, FFN cost (linear in tokens) dominates at current model scales. The probe stage runs at reduced spatial resolution ($224{\times}224$ in our experiments), making each probe forward markedly cheaper than a full-resolution pass. So the $2K$ probe passes plus a focused pass on $M_{\mathrm{eff}}$ full-resolution frames remain net-cheaper than a single full-resolution pass on all $N$ frames (Fig.~\ref{fig:pareto}).

\begin{figure}[t]
  \centering
\includegraphics[width=\linewidth]{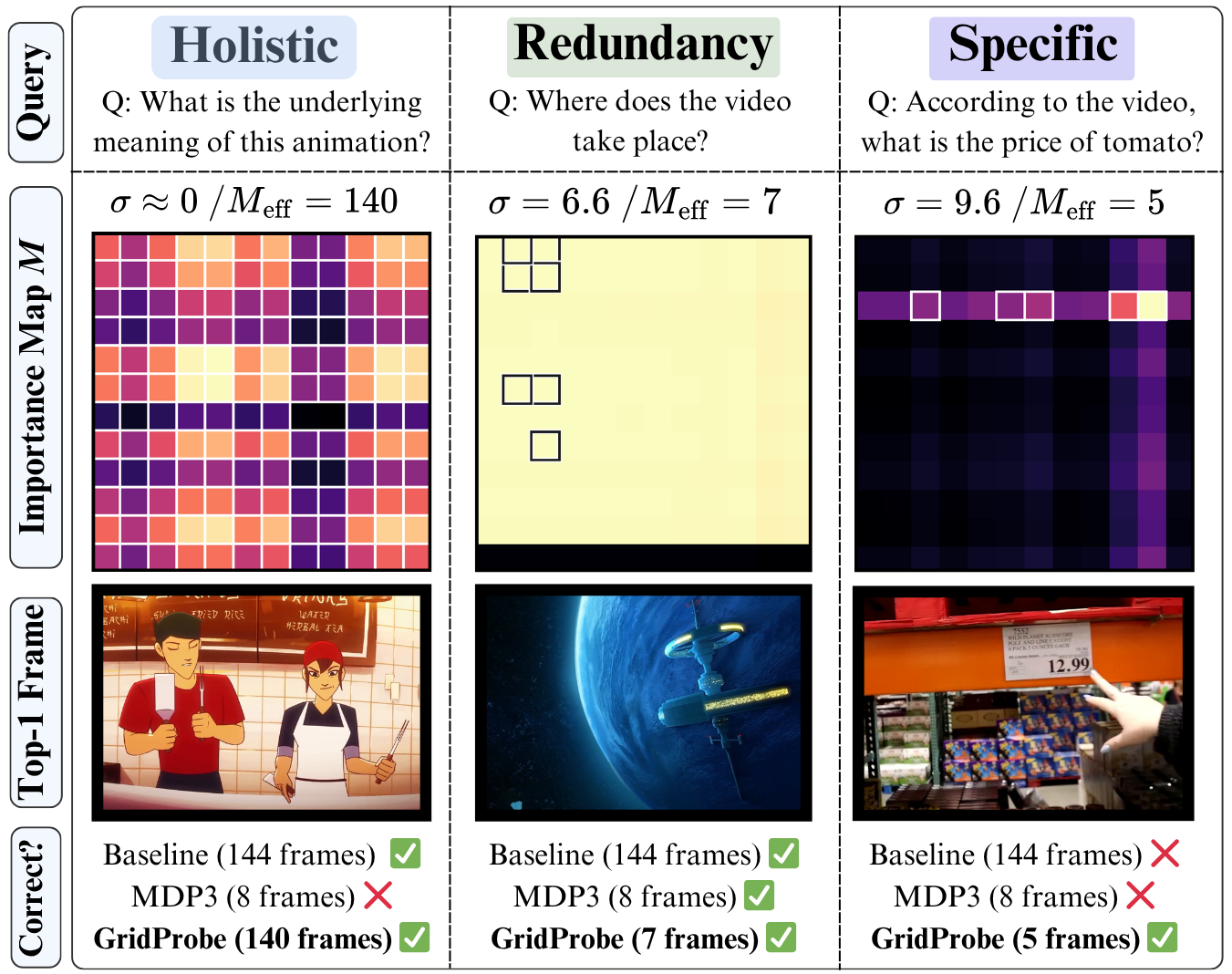}\caption{Three Video-MME-v2 queries exercise three distribution-shape regimes ($K{=}12$). The $\sigma$ statistic adapts $M_{\mathrm{eff}}$ from $140$ (holistic) to $5$ (specific). On the specific query, \method{} answers correctly with $5$ frames while the $K^2{=}144$ baseline fails. Notably, MDP3 (a powerful encoder-space selector with its paper-default fixed budget of $M{=}8$) misses both the holistic and specific cases.}  \label{fig:qmaps}
\end{figure}

\section{Experiments}\label{sec:experiments}

\subsection{Experimental Setup}

We evaluated on \textbf{Video-MME-v2}~\cite{fu2026video} (8-option MCQ, $3{,}200$ questions across 800 videos with three-level cognitive hierarchy and grouped non-linear scoring, reported visual-only with no subtitles) and \textbf{LongVideoBench}~\cite{wu2024longvideobench} (with subtitles). All backbones are Qwen3-VL-Instruct~\cite{bai2025qwen3} (2B, 4B, 8B), frozen at inference. Unless stated otherwise, $K{=}12$ (a $12{\times}12$ grid yielding a $144$-frame candidate pool), $\gamma_0{=}0.25$ in Eq.~\ref{eq:selector}, probe resolution $224{\times}224$ pixels, and uncapped focused-pass resolution. Frame sampling draws $K^2$ frames uniformly from the video timeline. We reported Average Accuracy, the official Non-Linear grouped score (VMME-V2 only), and per-question TFLOPs.

\subsection{Main Results: Video-MME-v2 and LongVideoBench}\label{sec:mainresults}

\begin{table}[t]
\centering
\caption{Main results on Video-MME-v2 (no subtitles) and LongVideoBench at $K{=}12$. \texttt{GP-X} denotes single-model \method{} (selector $=$ QA $=$ Qwen3-VL-X). \texttt{GP-X$\to$Y} denotes cross-model pipelines. Non-Lin is V2's official grouped score. Long is LVB's 3600-second bin. The Qwen3-VL-2B baseline is the comparison anchor for the lower blocks. \textcolor{gray!70}{numbers in gray} are for reference. \textbf{Bold} marks the best \method{} operating point per block.}
\label{tab:main}
\small
\setlength{\tabcolsep}{6pt}
\begin{tabular}{@{} l ccc ccc @{}}
\toprule
 & \multicolumn{3}{c}{\textbf{Video-MME-v2}} & \multicolumn{3}{c}{\textbf{LongVideoBench}} \\
\cmidrule(lr){2-4} \cmidrule(l){5-7}
Pipeline & Non-Lin & Avg Acc & TFLOPs & Long Acc & Overall & TFLOPs \\
\midrule
\multicolumn{7}{@{}l}{\emph{Monolithic Baselines (full $K^2{=}144$ pool). 2B is the comparison anchor.}} \\
\quad Qwen3-VL-2B            & 9.45  & 23.16 & 820  & 49.8 & 56.4 & 868  \\
\quad \textcolor{gray!70}{Qwen3-VL-4B}            & \textcolor{gray!70}{14.11} & \textcolor{gray!70}{30.06} & \textcolor{gray!70}{1415} & \textcolor{gray!70}{57.3} & \textcolor{gray!70}{64.1} & \textcolor{gray!70}{1493} \\
\quad \textcolor{gray!70}{Qwen3-VL-8B}            & \textcolor{gray!70}{14.91} & \textcolor{gray!70}{30.94} & \textcolor{gray!70}{2441} & \textcolor{gray!70}{55.1} & \textcolor{gray!70}{62.7} & \textcolor{gray!70}{2574} \\
\midrule
\multicolumn{7}{@{}l}{\emph{Single-Model \method{}: efficient trade-off at fixed model size (vs same-size baseline).}} \\
\quad GP-2B                  & 8.39  & 21.53 & 245  & 51.4 & 57.3 & 301  \\
\quad \textbf{GP-4B}         & \textbf{12.86} & \textbf{28.28} & \textbf{440}  & \textbf{55.9} & \textbf{62.4} & \textbf{575}  \\
\quad GP-8B                  & 12.60 & 28.06 & 842  & 53.4 & 60.7 & 1068 \\
\midrule
\multicolumn{7}{@{}l}{\emph{Cross-Model \method{}: Pareto-dominance over the 2B baseline (2B probe $\to$ larger QA).}} \\
\quad \textcolor{gray!70}{Uniform-$M_{\mathrm{eff}}$ $\to$ 8B} & \textcolor{gray!70}{10.83} & \textcolor{gray!70}{26.22} & \textcolor{gray!70}{677} & \textcolor{gray!70}{49.8} & \textcolor{gray!70}{58.5} & \textcolor{gray!70}{735} \\
\quad GP-2B$\to$4B           & 10.76          & 25.22          & 399           & \textbf{54.3} & \textbf{60.4} & \textbf{452}           \\
\quad GP-2B$\to$8B           & \textbf{11.70} & \textbf{26.72} & \textbf{677}  & 52.0          & 59.7          & 735           \\
\bottomrule
\end{tabular}
\end{table}

\begin{table}[t]
\begin{minipage}[t]{0.49\linewidth}
\centering
\caption{Selector quality at fixed $M{=}8$, 2B QA. Both use the exact same per-question frame budget. Only the scoring selector differs.}
\label{tab:abl_selector}
\small
\setlength{\tabcolsep}{4pt}
\begin{tabular}{l c c}
\toprule
2B QA, $M{=}8$ & V2 NL\,/\,Acc & LVB Long\,/\,Ov \\
\midrule
$+$ MDP3                                 & 7.38\,/\,20.09 & 44.9\,/\,51.0 \\
$+$ \method{}                                          & 7.49\,/\,20.12 & 47.3\,/\,54.6 \\
\midrule
$\Delta$                                              & \textbf{$+$0.11\,/\,$+$0.03} & \textbf{$+$2.4\,/\,$+$3.6} \\
\bottomrule
\end{tabular}
\end{minipage}\hfill
\begin{minipage}[t]{0.49\linewidth}
\centering
\caption{Adaptive $M$ vs fixed $M{=}8$ within \method{}, 2B QA. Both variants have comparable compute ($\sim$240T V2. $\sim$300T LVB).}
\label{tab:abl_adaptiveM}
\small
\setlength{\tabcolsep}{4pt}
\begin{tabular}{l c c}
\toprule
2B QA, GP & V2 NL\,/\,Acc & LVB Long\,/\,Ov \\
\midrule
$M{=}8$                                       & 7.49\,/\,20.12 & 47.3\,/\,54.6 \\
$M{=}\textsc{auto}$              & 8.39\,/\,21.53 & 51.4\,/\,57.3 \\
\midrule
$\Delta$                                              & \textbf{$+$0.90\,/\,$+$1.41} & \textbf{$+$4.1\,/\,$+$2.7} \\
\bottomrule
\end{tabular}
\end{minipage}
\end{table}


\textbf{Same-model results.} On Qwen3-VL-2B at $K{=}12$ (Block~2 of Table~\ref{tab:main}), \method{}($M{=}\textsc{auto}$) trades $1.63$ pp Avg Acc on V2 for a $3.36\times$ TFLOPs reduction, and reaches a Pareto-dominant point on LVB ($+0.9$ pp at $0.35\times$ compute). The trade-off is broadly invariant across QA size: comparing each GP-X to its same-size monolithic baseline, the accuracy cost is $-1.63$/$-1.78$/$-2.88$ pp on V2 at 2B/4B/8B and $-1.7$/$-2.0$ pp on LVB at 4B/8B, shifting modestly toward higher accuracy cost at larger backbones because stronger QAs extract more from the full-frame baseline. As a side benefit, GP-8B reaches $28.06\%$ Avg Acc on V2 at $842$ TFLOPs, matching the 2B baseline's compute (820 TFLOPs) within $3\%$ at $+4.9$ pp accuracy, a near-matched-compute upgrade for users willing to deploy the 8B answerer. 

\textbf{Cross-model: a free Pareto move.}\label{sec:crossmodel}
Pairing the 2B selector with a stronger QA (Block~3 of Table~\ref{tab:main}, visualized in Fig.~\ref{fig:pareto}(a)) Pareto-dominates the 2B-monolithic baseline on both benchmarks: $+3.56$ pp Avg Acc at $0.83\times$ compute on V2 and $+3.30$ pp at $0.85\times$ on LVB for GP-2B$\to$8B, with GP-2B$\to$4B delivering an even larger LVB gain ($+4.0$ pp at $0.52\times$, widening to $+4.5$ pp on the 3600-sec bin). The mechanism is straightforward: attention on $M_{\mathrm{eff}}$ frames at 8B is cheaper than on $K^2$ frames at 2B because sequence length dominates parameter count in the multi-frame regime, and the larger QA produces sharper answer posteriors on the focused subset. Most of the cross-model win comes from the adaptive sizing decision and the larger QA. The Uniform-$M_{\mathrm{eff}}{\to}$8B control (gray row in Table~\ref{tab:main}) uses GridProbe's per-question $M_{\mathrm{eff}}$ but draws those frames \emph{uniformly} from the $K^2$ pool (no importance ranking). Against this matched-compute baseline, the importance ranking adds a smaller residual: $+0.50$ pp on V2 and $+1.20$ pp on LVB.

\paragraph{Decomposing the same-model gap.}
Tables~\ref{tab:abl_selector} and~\ref{tab:abl_adaptiveM} factor the gap into selector quality (vs MDP3 at matched $M{=}8$) and adaptive sizing (within \method{}). At fixed $M{=}8$, the two selectors are matched on V2 but diverge sharply on LVB, where \method{} wins by $+3.6/+2.4$ pp Overall / 3600s while MDP3 lands $-4.9$ pp \emph{below} the no-selection baseline. Encoder-space scoring is not just suboptimal there but actively harmful. Figure~\ref{fig:qmaps} illustrates the failure mode qualitatively: MDP3 misses both the holistic and specific cases that \method{} answers correctly. Since MDP3 itself dominates a broad suite of training-free selectors (CLIP-based scoring, scene-change detection, optical flow, Frame-Voyager~\cite{yu2024frame}, and others)~\cite{sun2025mdp3}, the gap extends transitively over that family. Switching from fixed $M{=}8$ to $M{=}\textsc{auto}$ adds $+0.90 / +4.1$ Non-Lin / Long-bin pp at near-matched compute, isolating the contribution of $\sigma$-driven per-question allocation on top of the selector. The two effects together account for the gaps in Table~\ref{tab:main}.


\subsection{Adaptive Compute Mirrors Intrinsic Question Difficulty}\label{sec:adaptive_compute}

Figure~\ref{fig:skew} provides direct empirical evidence that \method{}'s adaptive selection responds to question content. \textbf{Firstly}, the $M_{\mathrm{eff}}$ curve is symmetric in skew sign: both extremes of the importance-map skewness axis route to small $M_{\mathrm{eff}}$, while the near-uniform middle gets near-full coverage. This is the redundancy principle of \S\ref{sec:selector} made empirical: positive-skew (sparse peaks) and negative-skew (redundant high-importance frames) are different distribution shapes with the \emph{same} selection answer. \textbf{Secondly}, baseline accuracy mirrors the $M_{\mathrm{eff}}$ curve: questions in the near-uniform regime are intrinsically harder ($\sim 21\%$ baseline accuracy) while questions at either tail are easier ($\sim 28$ to $32\%$), so the two curves visibly mirror each other across $\mathrm{skew}(M)$. The selector's compute allocation tracks intrinsic difficulty without ever observing the answer. \textbf{Thirdly}, the adaptive variability is quantitatively large: the cross-question coefficient of variation (CV, standard deviation over mean) in compute is $0.78$ for \method{} versus $0.018$ for the fixed-input baseline ($44\times$ higher), at $0.30\times$ the per-question average compute. This level of input-dependent compute variability is a signature of adaptive test-time compute that no fixed-input baseline can produce.


\section{Ablation Study}\label{sec:ablations}

We run three ablations: selector size in cross-model pairings, the temporal vs.\ importance order of the focused pass, and an image-collation efficiency variant. All ablations use Qwen3-VL-2B at $K{=}12$ on Video-MME-v2 with $M{=}\textsc{auto}$ unless stated.

\textbf{Selector size.}
Holding the QA fixed at 2B, scaling the selector from 2B to 8B \emph{decreases} Avg Acc while increasing TFLOPs (Table~\ref{tab:ablation_selector}), the opposite of the naive capability-scaling expectation.The per-question average $\bar{M}_{\mathrm{eff}}$ is roughly stable across selector sizes ($52.9 \to 56.2 \to 58.5$ for 2B/4B/8B), which suggests that the accuracy degradation is not driven by more aggressive selection. Instead, a stronger selector likely identifies frames informative for its own reasoning capacity, which need not align with what the smaller QA needs to answer. The cross-model amortization (\S\ref{sec:crossmodel}) is therefore one-directional: a small probe paired with a more capable answerer, not the reverse.

\textbf{Frame ordering.}
The focused pass receives the top-$M_{\mathrm{eff}}$ frames in temporal order. Passing them in descending-importance order (same frames, different positional encoding) costs $-1.25$ pp Avg Acc (Table~\ref{tab:ablation_order}). The effect is small but consistent: temporal ordering preserves the positional encoding the VLM was trained to read.

\begin{table}[t]
\begin{minipage}[t]{0.48\linewidth}
\centering
\caption{Selector-size ablation. VMME-V2 stratified $n{=}400$. $\bar{M}_{\mathrm{eff}}$ is the per-question average.}
\label{tab:ablation_selector}
\small
\setlength{\tabcolsep}{6pt}
\begin{tabular}{l l c c c}
\toprule
Selector & QA & Avg Acc & $\bar{M}_{\mathrm{eff}}$ & TFLOPs \\
\midrule
2B & 2B & \textbf{26.25}\% & \textbf{52.9} & \textbf{233.3} \\
4B & 2B & 26.00\% & 56.2 & 302.0 \\
8B & 2B & 24.75\% & 58.5 & 415.1 \\
\bottomrule
\end{tabular}
\end{minipage}\hfill
\begin{minipage}[t]{0.49\linewidth}
\centering
\caption{Frame-order ablation. V2 stratified $n{=}400$. Same frames, different input order.}
\label{tab:ablation_order}
\small
\setlength{\tabcolsep}{6pt}
\begin{tabular}{l c c}
\toprule
Frame order & Avg Acc & $\Delta$ \\
\midrule
\textbf{Temporal} (default) & \textbf{26.25\%} & (ref.) \\
Importance (descending) & 25.00\% & $-1.25$ pp \\
\bottomrule
\end{tabular}
\end{minipage}
\end{table}

\begin{table}[t]
\centering
\caption{Collated single-image variant vs the standard $M_{\mathrm{eff}}$-frame two-stage, V2 $n{=}1{,}800$.}
\label{tab:ablation_collated}
\small
\setlength{\tabcolsep}{10pt}
\begin{tabular}{l c c}
\toprule
Method & Avg Acc & TFLOPs \\
\midrule
Baseline (no selection)                                                       & 23.10          & 819.6        \\
Two-stage ($M_{\mathrm{eff}}$ frames as video)                                & \textbf{21.27} & 232.2        \\
Collated (one $\sqrt{M_{\mathrm{eff}}}{\times}\sqrt{M_{\mathrm{eff}}}$ image) & 20.11          & \textbf{66.6} \\
\bottomrule
\end{tabular}
\end{table}

\textbf{Image collation.}
As an efficiency variant, the focused pass can composite the top-$M_{\mathrm{eff}}$ frames into a single $\lceil\sqrt{M_{\mathrm{eff}}}\rceil{\times}\lceil\sqrt{M_{\mathrm{eff}}}\rceil$ tiled image at $2048{\times}2048$ (with empty cells when $M_{\mathrm{eff}}$ is not a perfect square), reducing compute to one image's worth of tokens at the cost of temporal positional encoding and per-frame pixel budget. Collation reaches $0.29\times$ the standard two-stage's compute at a $-1.16$ pp Avg Acc cost (Table~\ref{tab:ablation_collated}), making it a credible operating point when extreme compute is the constraint.

\section{Conclusion and Limitations}\label{sec:conclusion}\label{sec:limitations}

We introduced \method{}, a training-free posterior-probing paradigm in which row and column probes over a $K{\times}K$ grid produce a question-conditioned importance map, and a closed-form shape statistic sets the per-question budget $M_{\mathrm{eff}}$. \method{} delivers Pareto-efficient single-model operation on Video-MME-v2 and Pareto-dominant operation on LongVideoBench, the cross-model variant (2B selector with a stronger QA) Pareto-dominates the 2B-monolithic baseline on both benchmarks without retraining, and the per-question $M_{\mathrm{eff}}$ tracks intrinsic question difficulty without ever observing the answer.

A few caveats and natural refinements remain. The TFLOPs reduction is most pronounced when the focused pass dominates inference, and more modest with large prompts (e.g., LVB's $\sim$700--1{,}000 subtitle tokens that every probe re-processes) or small grid sizes ($K{<}10$, where the $2K$ probe passes themselves become a non-trivial fraction of total cost). The cross-model pipeline shifts cost from compute toward host memory, since both selector and QA are loaded simultaneously. Two refinements are natural follow-ups: allowing $\gamma_0$ to adapt to video length or pool density, and extending the shape statistic to other per-cell importance signals such as attention magnitudes or retrieval scores. Finally, our probe confidence $\max_y p_\theta(y \mid S, q)$ is defined for the finite answer space of multiple-choice benchmarks. Generalization to open-ended QA is non-trivial and left to future work.

\section{Acknowledgment}
We are grateful to the KAUST Academy for its generous support, and especially to Prof. Sultan Albarakati who
made this work possible. For computer time, this research used Ibex managed by the Supercomputing Core Laboratory at King Abdullah University of Science \& Technology (KAUST) in Thuwal, Saudi Arabia.

\small
\bibliographystyle{unsrt}
\bibliography{references}

\clearpage

\appendix
\section{Implementation Details}\label{sec:app-impl}

\paragraph{Models.} All experiments use Qwen3-VL-Instruct backbones at three sizes (2B, 4B, 8B parameters) loaded from HuggingFace Transformers. We perform zero-shot inference: no fine-tuning, no chain-of-thought prompting, no tool-use. Final answers are read off via letter-token scoring, taking $\arg\max_y p_\theta(y \mid \cdot)$ over $y \in \{A, B, C, D, E, F, G, H\}$ (or the letter set for LongVideoBench's 5-way questions).

\paragraph{Hyperparameters.} The defaults used throughout the paper are $K{=}12$ (so $K^2{=}144$ frames in the pool), $\gamma_0{=}0.25$ for the $M_{\mathrm{eff}}$ closed-form rule, and a small variance threshold to set $\sigma{=}0$ in the degenerate near-uniform case (\S\ref{sec:selector}). The probe stage runs at reduced spatial resolution $224{\times}224$. the focused stage runs at the model's native resolution with Qwen3-VL's adaptive per-frame token allocation.

\paragraph{Pipeline.} For each (video, question) pair, we extract $K^2{=}144$ frames uniformly from the video duration. Stage 1 runs $2K{=}24$ probe passes through the frozen VLM ($K$ row passes and $K$ column passes), where each pass collates $K{=}12$ frames into a single $\lceil\sqrt{K^2}\rceil{\times}\lceil\sqrt{K^2}\rceil$ tiled image at $2048{\times}2048$. Each pass produces a softmax confidence over the answer letters; we take the top-1 confidence as the probe score per row/column. The importance map is $M[r,c] = c_r \cdot c_c$ (Eq.~\ref{eq:map}). We then compute the shape statistic $\sigma = |\mathrm{skew}(M)| + \frac{1}{2}\max(0, \mathrm{kurt_{ex}}(M))$ and the per-question budget $M_{\mathrm{eff}} = K^2 / (1 + \gamma_0 K \sigma)$. Stage 2 runs a single forward pass on the top-$M_{\mathrm{eff}}$ frames (by $M$ score) at the model's native resolution.

\paragraph{Benchmarks.} \textbf{Video-MME-v2} (visual-only): $n{=}3{,}200$ questions across 800 videos in 4 duration bins (15s, 60s, 600s, 3600s). We report Non-Lin (V2's official grouped score) and Avg Acc. \textbf{LongVideoBench validation set} (with subtitles): subtitles are concatenated to the prompt as text in both the probe and focused stages. We report Long (3600-second bin) and Overall accuracy.

\paragraph{Hardware.} All experiments ran on a single node with $8{\times}$ NVIDIA A100 GPUs. We shard each evaluation across the 8 GPUs by sample id (interleaved partitioning to balance the duration-bin distribution across shards) and merge per-shard JSON outputs. Source code will be released publicly upon publication.

\section{Detailed Breakdown Tables}\label{sec:app-tables}

This section provides the per-bin breakdown referenced from Table~\ref{tab:main}.

\subsection{Video-MME-v2: Per-Level and per-Group-Type}\label{sec:app-vmme-breakdown}

Video-MME-v2 organizes questions into three cognitive levels (L1: Information Aggregation, L2: Temporal Understanding, L3: Complex Reasoning) and two group types (\emph{Consistency} for capability-consistency groups, \emph{Coherence} for reasoning-coherence groups). The Consistency block contains 519 of 800 groups and the Coherence block contains 281 of 800 groups. Table~\ref{tab:breakdown} reports Non-Lin score for each method at each breakdown axis.

Against the same-model 2B baseline, \method{}($M{=}\textsc{auto}$) \emph{wins} on Level~2 temporal understanding ($\Delta{=}{+}0.37$ Non-Lin), where evidence is concentrated at a transition-bearing event, and \emph{loses} most on Level~1 multi-point aggregation ($\Delta{=}{-}1.92$ Non-Lin), where evidence is genuinely distributed across many timestamps and no small subset suffices. The $\sigma$-driven adaptive $M$ degrades gracefully toward $K^2$ on near-uniform maps, so the method does not overcommit to selection when evidence is dispersed but has nothing to gain when it is already fully covered.

\begin{table}[h]
\centering
\caption{V2 Non-Lin breakdown by cognitive level (L1/L2/L3) and group type (Consistency/Coherence). Visual-only, $K{=}12$, $n{=}3{,}200$.}
\label{tab:breakdown}
\small
\setlength{\tabcolsep}{4pt}
\begin{tabular}{l ccc cc c}
\toprule
Method & Level 1 & Level 2 & Level 3 & Consistency & Coherence & Overall \\
\midrule
\multicolumn{7}{l}{\emph{Single-model: Qwen3-VL-2B-Instruct}} \\
Baseline (no selection)                  & 13.83 & 7.50  & 8.30  & 10.18 & 8.12  & 9.45 \\
\quad + \method{} ($M{=}$\textsc{auto})  & 11.91 & 7.87  & 6.72  & 9.13  & 7.03  & 8.39 \\
\midrule
\multicolumn{7}{l}{\emph{Same-model: Qwen3-VL-4B-Instruct}} \\
Baseline (no selection)                  & 19.43 & 12.56 & 12.12 & 15.04 & 12.40 & 14.11 \\
\quad + \method{} ($M{=}$\textsc{auto})  & 18.98 & 10.73 & 10.82 & 14.20 & 10.40 & 12.86 \\
\midrule
\multicolumn{7}{l}{\emph{Same-model: Qwen3-VL-8B-Instruct}} \\
Baseline (no selection)                  & 18.98 & 15.19 & 12.32 & 15.53 & 13.75 & 14.91 \\
\quad + \method{} ($M{=}$\textsc{auto})  & 16.18 & 11.78 & 11.11 & 12.95 & 11.97 & 12.60 \\
\midrule
\multicolumn{7}{l}{\emph{Cross-model: 2B selector $\to$ 4B QA}} \\
4B baseline (no selection)               & 19.43 & 12.56 & 12.12 & 15.04 & 12.40 & 14.11 \\
\quad + \method{} (2B selector, ours)    & 14.42 & 9.30  & 9.68  & 11.37 & 9.64  & 10.76 \\
\midrule
\multicolumn{7}{l}{\emph{Cross-model: 2B selector $\to$ 8B QA}} \\
8B baseline (no selection)               & 18.98 & 15.19 & 12.32 & 15.53 & 13.75 & 14.91 \\
\quad + \method{} (2B selector, ours)    & 15.28 & 11.26 & 9.93  & 12.02 & 11.12 & 11.70 \\
\bottomrule
\end{tabular}
\end{table}

\subsection{LongVideoBench: Full Per-Duration Breakdown}\label{sec:app-lvb-breakdown}

The main table reports only LongVideoBench's \emph{Long} bin and \emph{Overall}. Table~\ref{tab:lvb_full} adds the four duration buckets. \method{}($M{=}\textsc{auto}$) \emph{wins} on the 600-sec ($+2.2$ pp) and 3600-sec ($+1.6$ pp) bins, where uniform $K^2$ undersamples the timeline and question-conditioned selection picks the relevant moments, and \emph{loses} on the 15-sec and 60-sec bins ($-3.1$ pp each), where $144$ uniform frames already saturate the timeline and selection is unnecessary. The same pattern appears on V2's cognitive levels (\S\ref{sec:app-vmme-breakdown}): the method earns its keep where evidence is sparse and degrades gracefully when it is already fully covered.

\begin{table}[h]
\centering
\caption{LongVideoBench Avg Acc \% by source-video duration (15s / 60s / 600s / 3600s / Overall), $K{=}12$, with subtitles. $\Delta$ rows: \method{}'s gain over the corresponding no-selection baseline.}
\label{tab:lvb_full}
\small
\setlength{\tabcolsep}{6pt}
\begin{tabular}{l ccccc}
\toprule
Method & 15-sec & 60-sec & 600-sec & 3600-sec & Overall \\
\midrule
\multicolumn{6}{l}{\emph{Single-model: Qwen3-VL-2B-Instruct}} \\
2B baseline (no selection)               & 72.9 & 77.3 & 55.3 & 49.8 & 56.4 \\
\quad + \method{} ($M{=}8$, fixed)       & 66.7 & 69.1 & 57.3 & 47.3 & 54.6 \\
\quad + \method{} ($M{=}$\textsc{auto})  & 69.8 & 74.2 & \textbf{57.5} & \textbf{51.4} & \textbf{57.3} \\
\midrule
$\Delta$ \method{} (auto) vs.\ 2B base.   & $-3.1$ & $-3.1$ & $\mathbf{+2.2}$ & $\mathbf{+1.6}$ & $\mathbf{+0.9}$ \\
\midrule
\multicolumn{6}{l}{\emph{Same-model: Qwen3-VL-4B-Instruct (selector $=$ QA)}} \\
4B baseline (no selection)               & 81.4 & 79.4 & 64.6 & 57.3 & 64.1 \\
\quad + \method{} ($M{=}\textsc{auto}$, same-model) & 79.8 & \textbf{80.4} & 61.7 & 55.9 & 62.4 \\
\midrule
\multicolumn{6}{l}{\emph{Same-model: Qwen3-VL-8B-Instruct (selector $=$ QA)}} \\
8B baseline (no selection)               & 80.6 & 74.2 & 64.8 & 55.1 & 62.7 \\
\quad + Uniform sampling                 & 77.5 & 76.3 & 60.9 & 51.1 & 59.3 \\
\quad + \method{} ($M{=}\textsc{auto}$, same-model) & 76.7 & 74.2 & 62.6 & 53.4 & 60.7 \\
\midrule
\multicolumn{6}{l}{\emph{Cross-model: 2B selector $\to$ 4B QA}} \\
\quad + Uniform sampling                 & 77.5 & 78.4 & 60.0 & 50.4 & 58.8 \\
\quad + \method{} (2B selector, ours)    & 76.7 & 75.3 & 60.2 & 54.3 & \textbf{60.4} \\
\midrule
$\Delta$ \method{} (2B$\to$4B) vs.\ 2B base. & $\mathbf{+3.8}$ & $-2.0$ & $\mathbf{+4.9}$ & $\mathbf{+4.5}$ & $\mathbf{+4.0}$ \\
\midrule
\multicolumn{6}{l}{\emph{Cross-model: 2B selector $\to$ 8B QA}} \\
\quad + Uniform sampling                 & 79.8 & 75.3 & 59.7 & 49.8 & 58.5 \\
\quad + \method{} (2B selector, ours)    & 76.0 & 74.2 & 61.7 & 52.0 & 59.7 \\
\midrule
$\Delta$ \method{} (2B$\to$8B) vs.\ 2B base. & $\mathbf{+3.1}$ & $-3.1$ & $\mathbf{+6.4}$ & $\mathbf{+2.2}$ & $\mathbf{+3.3}$ \\
\bottomrule
\end{tabular}
\end{table}

\clearpage
\end{document}